\documentclass[runningheads]{llncs}
\usepackage{graphicx}
\usepackage{xcolor} 
\usepackage{amssymb,amsmath,array}

\usepackage{cleveref}

\usepackage{graphicx}
\usepackage{subfigure}
\usepackage{caption}
\begin{document}
\title{Correlation visualization under missing values: \\a comparison between imputation and direct parameter estimation methods}

\author{Nhat-Hao Pham$^{1,2,*}$, Khanh-Linh Vo$^{1,2,*}$, Mai Anh Vu$^{1,2}$,  Thu Nguyen$^3$, \\ Michael A. Riegler$^3$, P{\aa}l Halvorsen$^3$, and Binh T. Nguyen$^{1,2}$ 
%
\vspace{.3cm}\\
%
$^1$University of Science - Faculty of Mathematics and Computer Science  \\
Ho Chi Minh City, Vietnam
\vspace{.1cm}\\
$^2$ Vietnam National University Ho Chi Minh City, Vietnam
%
\vspace{.1cm}\\
$^3$SimulaMet - Dept. of Holistic Systems,
Oslo, Norway\\
\footnote{denotes equal contribution}
}

\author{Nhat-Hao Pham$^\star$\inst{ 1,2} \and
 Khanh-Linh Vo$^\star$\inst{1,2} \and
Mai Anh Vu \thanks{denotes equal contribution}\inst{ 1,2} 
\and Thu Nguyen \inst{3} \and Michael A. Riegler \inst{3}, P{\aa}l Halvorsen \inst{3} \and Binh T. Nguyen \inst{1,2}}
\authorrunning{Mai Anh Vu et al.}

\institute{University of Science - Faculty of Mathematics and Computer Science  \\
Ho Chi Minh City, Vietnam \and
Vietnam National University Ho Chi Minh City, Vietnam \and 
SimulaMet, Oslo, Norway 
}
\maketitle              


\begin{abstract}


Correlation matrix visualization is essential for understanding the relationships between variables in a dataset, but missing data can seriously affect this important data visualization tool. In this paper, we compare the effects of various missing data methods on the correlation plot, focusing on two randomly missing data and monotone missing data. 
We aim to provide practical strategies and recommendations for researchers and practitioners in creating and analyzing the correlation plot under missing data. Our experimental results suggest that while imputation is commonly used for missing data, using imputed data for plotting the correlation matrix may lead to a significantly misleading inference of the relation between the features. In addition, the most accurate technique for computing a correlation matrix (in terms of RMSE) does not always give the correlation plots that most resemble the one based on complete data (the ground truth).  We recommend using DPER \cite{NGUYEN2022108082}, a direct parameter estimation approach, for plotting the correlation matrix based on its performance in the experiments. 

\end{abstract} 


\section{Introduction}\label{intro-relate}
A correlation plot is a powerful and important visualization tool to summarise datasets, reveal relationships between variables, and identify patterns in the given dataset. However, missing data is a common situation in practice. 
There are two main approaches to computing correlation when the data contains missing entries. 
The first is to impute missing data and then estimate the correlation from imputed data. The second way is to estimate the correlation matrix directly from missing data. 

Various imputation methods have been proposed to handle missing data. 
For example, some traditional methods that have been used for missing data include mean imputation and mode imputation. In addition, multivariate imputation by chained equations (MICE) \cite{buuren2010mice} is an iterative approach that can provide the uncertainty of the estimates.
Furthermore, {machine learning} techniques train on the available data to learn patterns and relationships and then are used to predict the missing values. There are many typical approaches, including tree-based models \cite{stekhoven2012missforest}, matrix completion \cite{mazumder2010spectral}, or clustering-based methods such as KNN Imputation (KNNI). 
Finally, {deep learning} is an emerging trend for handling missing data, due to the excellent prediction of deep learning architectures. Notable deep learning methods for imputation can be Generative
Adversarial Imputation Nets (GAIN) \cite{yoon2018gain} and Graph Imputer Neural Network (GINN) \cite{Indro2020ginn}.
These methods have achieved high accuracy but require significant computational resources and are best suited for data-rich cases.

For the direct parameter estimate approach, a simple way to estimate the covariance matrix from the data is to use case deletion. 
However, the accuracy of such an approach may be low. Recently, direct parameter estimation approaches were proposed to estimate the parameters from the data directly. For example, the EPEM algorithm \cite{NGUYEN20211} estimates these parameters from the data by maximum likelihood estimation for monotone missing data. Next, the DPER algorithm \cite{NGUYEN2022108082} is based on maximum likelihood estimation for randomly missing data \cite{NGUYEN2022108082}. The computation for the terms in the covariance matrix is based on each pair of features. Next, the PMF algorithm \cite{nguyen4260235pmf} for missing dataset with many observed features by using principle component analysis on the portion of observed features and then using maximum likelihood estimation to estimate the parameters. Such algorithms are scalable and have promising implementation speed. 


Since the correlation plot is an important tool in data analysis, it is necessary to understand the impact of missing data on the correlation plot
\cite{kraus2020assessing}. This motivates us to study the effects of various imputation and parameter estimation techniques on the correlation plot. 
In summary, our contributions are as follows. First, we examine the effect of different imputation and direct parameter estimation methods on the correlation matrix and the correlation matrix heat maps for randomly missing data and monotone missing data. Second, we show that imputation techniques that yield the lowest RMSE may not produce the correlation plots that most resemble the one based on completed data (the ground truth). Third, we illustrate that relying solely on RMSE to choose an imputation method and then plotting the correlation heat map based on the imputed data from that imputation method can be misleading. Last but not least, our analysis process using the Local RMSE differences heatmaps and Local difference (matrix subtraction) for correlation provides practitioners a strategy to draw inferences from correlation plots  under missing data.

\section{Methods Under Comparison} \label{sec-methods}

This section details techniques under comparison.  These include some state-of-the-art techniques (DPER, GAIN, GINN, SoftImpute, ImputePCA, missForest), and some traditional but remaining widely used methods (KNNI, MICE, Mean imputation). More details are as follows, (Except for DPER \cite{NGUYEN2022108082}, which directly estimates the covariance matrix from the data, the remaining methods involve imputing missing values prior to calculating the covariance matrix.)
\begin{itemize}
    \item \textbf{Mean imputation} imputes the missing entries for each feature by the mean of the observed entries in the corresponding feature.  


\item \textbf{Multiple Imputation by Chained Equations (MICE) \cite{scikit-learn}} 
 works by using a series of regression models that fit the observed data. In each iteration of the process, the missing values for a specific feature are imputed based on the other features in the dataset. 



\item  \textbf{MissForest \cite{stekhoven2012missforest}} is an iterative imputation method based on a random forest. It averages over many unpruned regression or classification trees, allowing a multiple imputation scheme and allowing estimation of imputation error.

\item \textbf{K-nearest neighbor imputation (KNNI)} 
is based on the K-nearest neighbor algorithm. For each sample to be imputed, KNNI finds a set of K-nearest neighbors and imputes the missing values with a mean of the neighbors for continuous data or the mode for categorical data.

\item \textbf{SoftImpute \cite{mazumder2010spectral}} is an efficient algorithm for large matrix factorization and completion. It uses convex relaxation to produce a sequence of regularized low-rank solutions for completing a large matrix. 


\item \textbf{ImputePCA (iterative PCA) \cite{josse2016missmda}} minimizes the criterion, and the imputation of missing values is achieved during the estimation process. Since this algorithm is based on the PCA model, it takes into account the similarities between individuals and the relationships between variables.  

\item \textbf{Expectation Maximization algorithm (EM)}  \cite{em1977dempster} 
is a maximum likelihood approach from incomplete data. It iteratively imputes missing values and updates the model parameters until convergence. The algorithm consists of two steps: E-step estimates missing data given observed data, and M-step estimates model parameters given complete data. 

\item \textbf{Generative Adversarial Imputation Networks (GAIN) \cite{yoon2018gain}}  adapts the GAN architecture where the generator imputes the missing data, while the discriminator distinguishes between observed and imputed data. The generator maximizes the discriminator's error rate, while the discriminator minimizes classification loss. 


\item \textbf{Graph Imputation Neural Networks (GINN) \cite{Indro2020ginn}} is an autoencoder architecture that completes graphs with missing values. It consists of a graph encoder and a graph decoder, which utilize the graph structure to propagate information between connected nodes for effective imputation. 


\item \textbf{Direct Parameter Estimation for Randomly missing data (DPER algorithm) \cite{NGUYEN2022108082}} is a recently introduced algorithm that directly finds the maximum likelihood estimates for an arbitrary one-class/multiple-class randomly missing dataset based on some mild assumptions of the data. Since the correlation plot does not consider the label, the DPER algorithm used for the experiments is the DPER algorithm for one class dataset.
\end{itemize}

\section{Experiments} 

\label{sec-experiments} 





\subsection{Experiment Settings} 

This section details the experimental setup for comparing the effects of the methods presented in Section \ref{sec-methods} on correlation plots with missing data. 
For the experiments on randomly missing data, we used the Iris and Digits datasets and generated randomly missing data with rates from $10\%$ to $50\%$. Here, the missing rate is the ratio between the number of missing entries and the total number of entries. For the monotone missing pattern, we removed a $50\%$ piece of the image in the right corner with $40\%, 50\%$, and $60\%$ height and width in the MNIST dataset \cite{lecun1998mnist}. For MNIST, as KNNI and GINN \cite{Indro2020ginn} cannot produce the results within 3 hours of running, we exclude them from the plot of results. For each dataset, we normalize the data and
run the experiments on a Linux machine with 16 GB of RAM and 4 cores, Intel i5-7200U, 3.100GHz. 
\begin{table}[!t]
  \centering
  \begin{tabular}{|c|c|c|c|}
    \hline
    Dataset     & \# Classes     & \# Features & \# Samples \\\hline
    Iris & 3 & 4 & 150\\
   
    Digits & 10 & 64 & 1797\\
    MNIST \cite{lecun1998mnist} & 10 & 784 & 70000 \\  \hline   
  \end{tabular}
  \vspace{1mm}
   \caption{The description of datasets used in the experiments.} 
   \label{table_info_datasets}
   \vspace{-10mm}
\end{table}


To throughout investigate the effects of missing data on the correlation heatmap, we employ three types of heatmap plots:  
\begin{itemize}
    \item \textbf{Correlation Heatmaps} illustrates the correlation matrix and employ a blue-to-white-to-red color gradient to illustrate correlation coefficients. The colormap utilizes distinct colors: blue, white, and red, which are represented by the values $[(0, 0, 1)$, $(1, 1, 1)$, $(1, 0, 0)]$ respectively, as defined by the color map of the \textit{matplotlib} library.
Blue indicates negative values, red denotes positive values, and white indicates median (0) values. The colormap's range is bounded within \textit{vmin = -1} and \textit{vmax = 1}, which are parameters of \textit{matplotlib} that linearly map the colors in the colormap from data values \textit{vmin}  to \textit{vmax}.
    
\item \textbf{Local RMSE Difference Heatmaps for Correlation} portray the RMSE differences on a cell-by-cell basis, comparing correlation matrices produced by each estimation technique with the ground truth correlation matrix.  We adopt a different color gradient, which is a green-to-white color map with values $[(1, 1, 1)$, $(0, 0.5, 0)]$, further enhancing the clarity of these visualizations. The intensity of the green colors corresponds to the magnitude of the difference between the correlation matrix and the ground truth (depicted as white).  

\item \textbf{Local Difference (Matrix Subtraction) Heatmaps for Correlation} show the local differences for a cell-by-cell comparison (matrix subtraction) between correlation matrices generated from each estimation technique and the ground truth correlation matrix. This plot shows whether the difference between the two plots is positive or negative, providing a clear indication of the direction of the difference.
\end{itemize}     
For each dataset, we provide five heatmap plots. The first two plots display the method outcomes across varying levels of missing value, with the first showing the Correlation Heatmaps and the second showing the Local RMSE Difference Heatmaps. The two subsequent plots provide a more detailed examination, specifically highlighting the highest missing rate of the Correlation Heatmaps and Local RMSE difference heatmaps. This enables a more thorough analysis of color variations that might have been challenging to detect in the first two plots due to spatial limitations. The final plot, also focusing on the highest missing rate, shows the Local Difference (Matrix Subtraction) Heatmaps for Correlation, clearly indicating the direction of the difference.   

We use gray to differentiate null values, especially when all cells in a feature share the same values, often near the image edges. This gray shade distinguishes null values from 0 values, which are depicted in white.
Furthermore, the plot arrangement follows a descending order of RMSE, focusing on the highest missing rate's RMSE due to its significant impact on color intensity across rates. To maintain consistency with the correlation plot, we represent the ground truth heatmaps difference in white.  
In addition, in the Local RMSE Difference Heatmaps for Correlation charts, we integrate the calculation of the RMSE, which represents the difference between the two correlation matrices while excluding null positions. 

Moreover, we include the dense rank in ascending order alongside it. Dense ranking is a method for giving ranks to a group of numbers. The ranks are given so that there are no gaps between them. The next rank isn't skipped if some numbers share the same rank. For instance, let's take this set of values: [89, 72, 72, 65, 94, 89, 72]. The resulting dense ranks would be: [1, 2, 2, 2, 3, 3, 4]. Below each Local RMSE Difference Heatmaps plot, we denote the rank and the RMSE value using the following structure: $(<\text{rank}>) \text{RMSE:} <\text{RMSE  value}>$. 



\subsection{Result \& Analysis}   
\begin{figure*}[b!]
    \centering

    \subfigure[]{
        \includegraphics[height=0.245\linewidth]{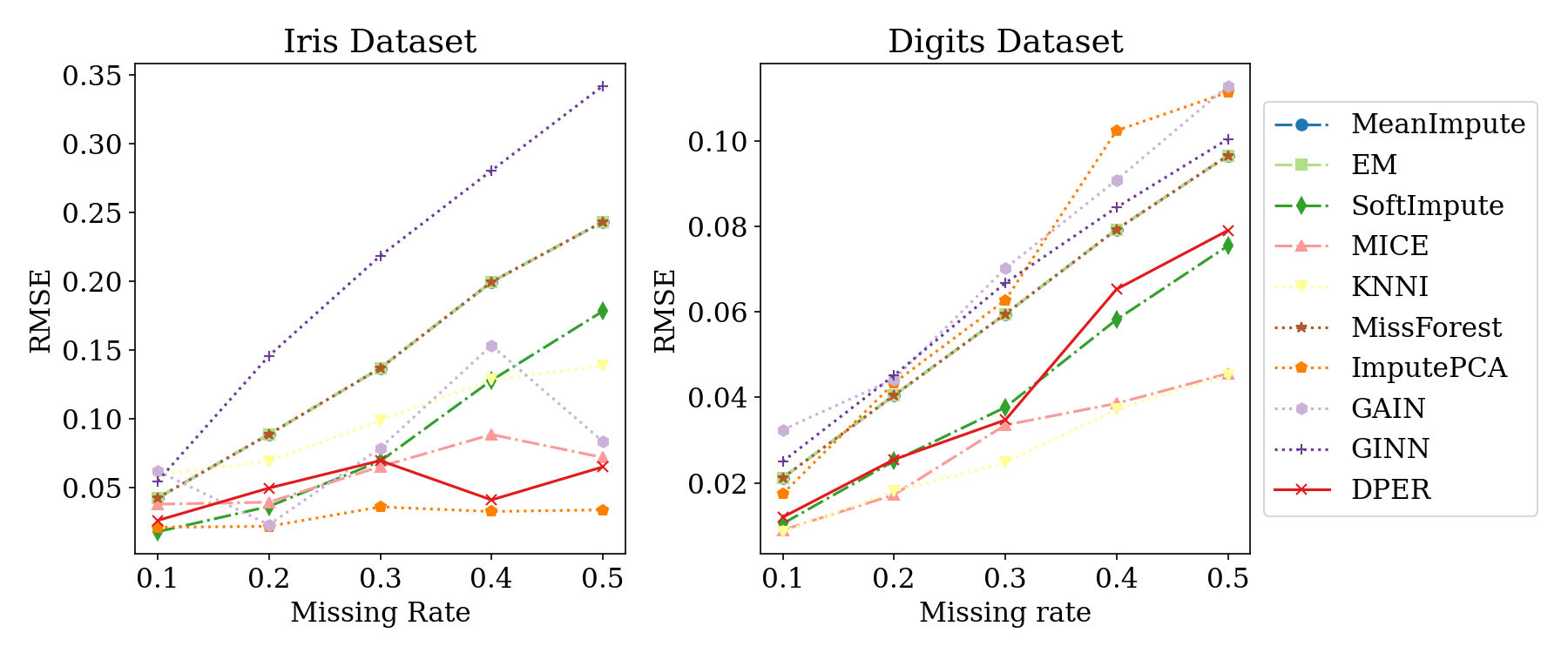}
        \label{subfig:line_1}
    }
    \subfigure[]{
        \includegraphics[height=0.245\linewidth]{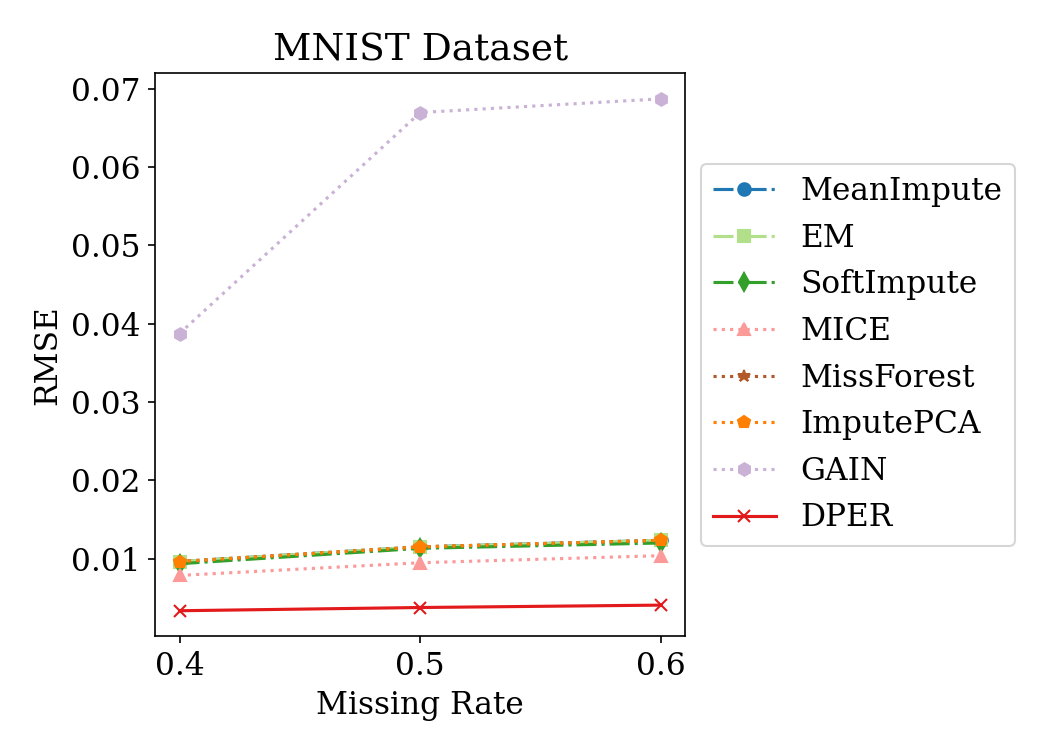}
        \label{subfig:line_2}
    }

    \caption{The RMSE difference of the correlation matrix between each method and the ground truth is depicted across missing rates ranging from $10\% \text{ to } 50\%$ for (a) Iris and Digits dataset with randomly missing patterns and (b) MNIST dataset with monotone missing patterns. }
    \label{fig:rmse_plots}
\end{figure*}
 
\subsubsection{Randomly missing data}  

\begin{figure*}[h]
    \centering
    \subfigure[Correlation Heatmaps] 
    {
        \label{subfig:iris_all_1}
        \includegraphics[width=\textwidth]{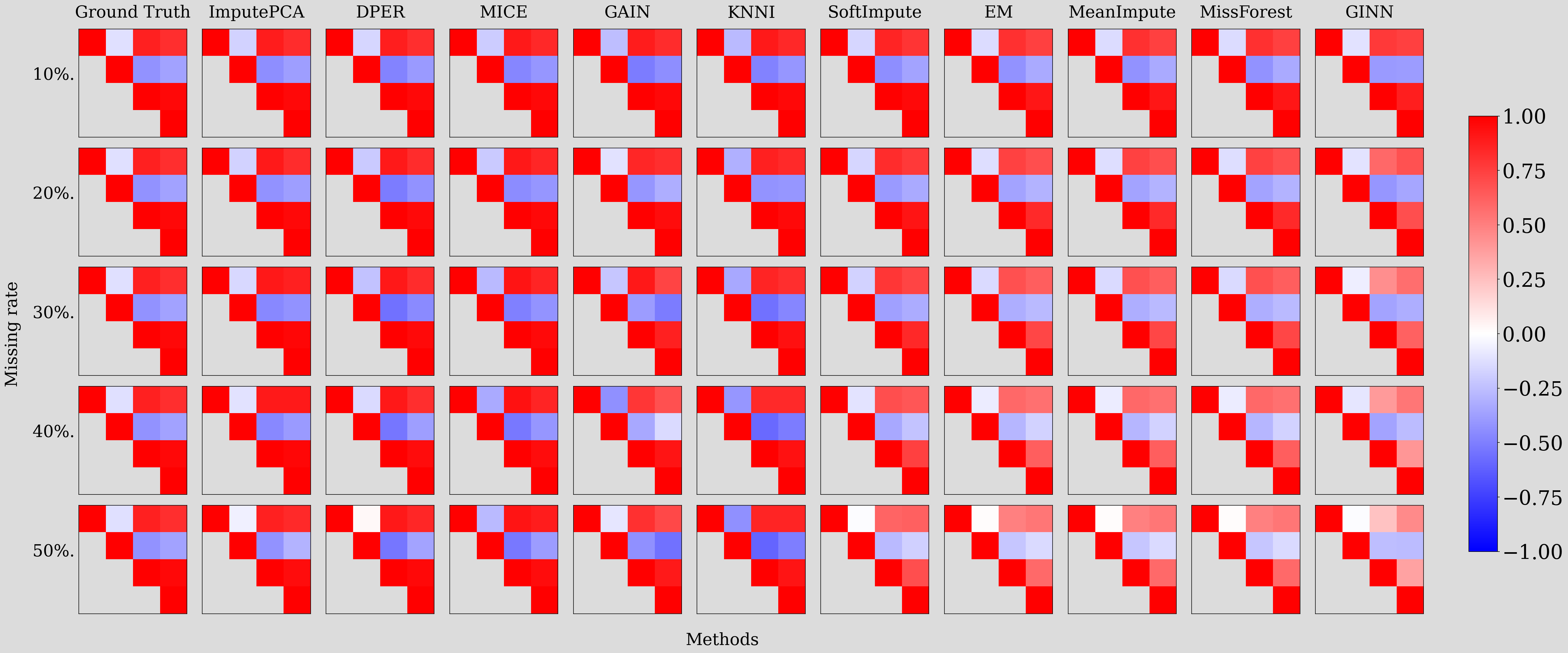}
    }\vspace{-2mm}\quad
    \subfigure[Local RMSE Difference Heatmaps for Correlation]
    {
        \label{subfig:iris_all_2}
        \includegraphics[width=\textwidth]{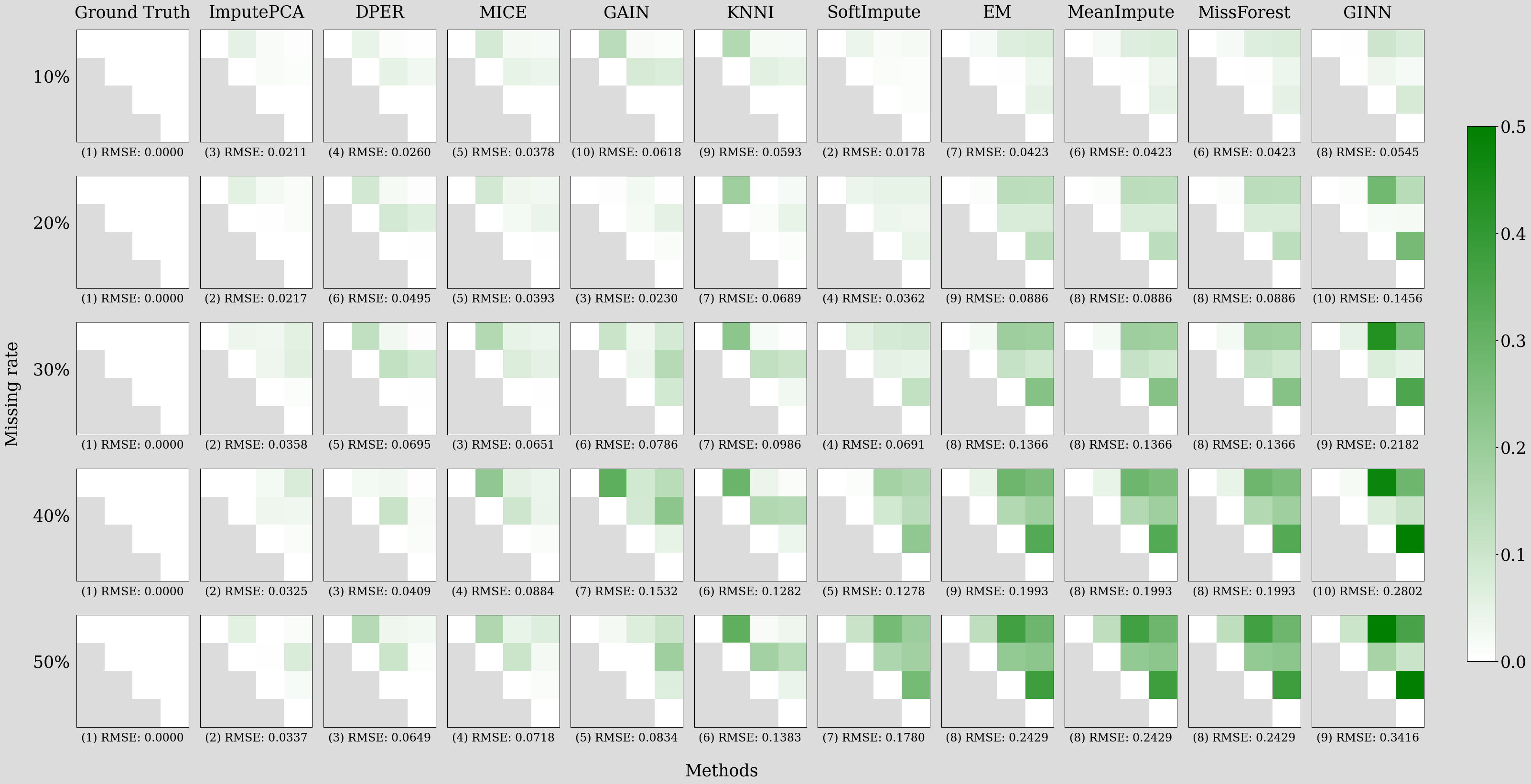}
    }\vspace{-2mm}\quad 
    \caption{Heatmaps for the Iris dataset across missing rates from $10\%$ to $50\%$}
    \label{fig:iris_all}
\end{figure*} 
  
\begin{figure*}[h!]
    \centering
    \subfigure[Correlation Heatmaps]
    {
        \label{subfig:iris_1}
        \includegraphics[width=0.9\textwidth]{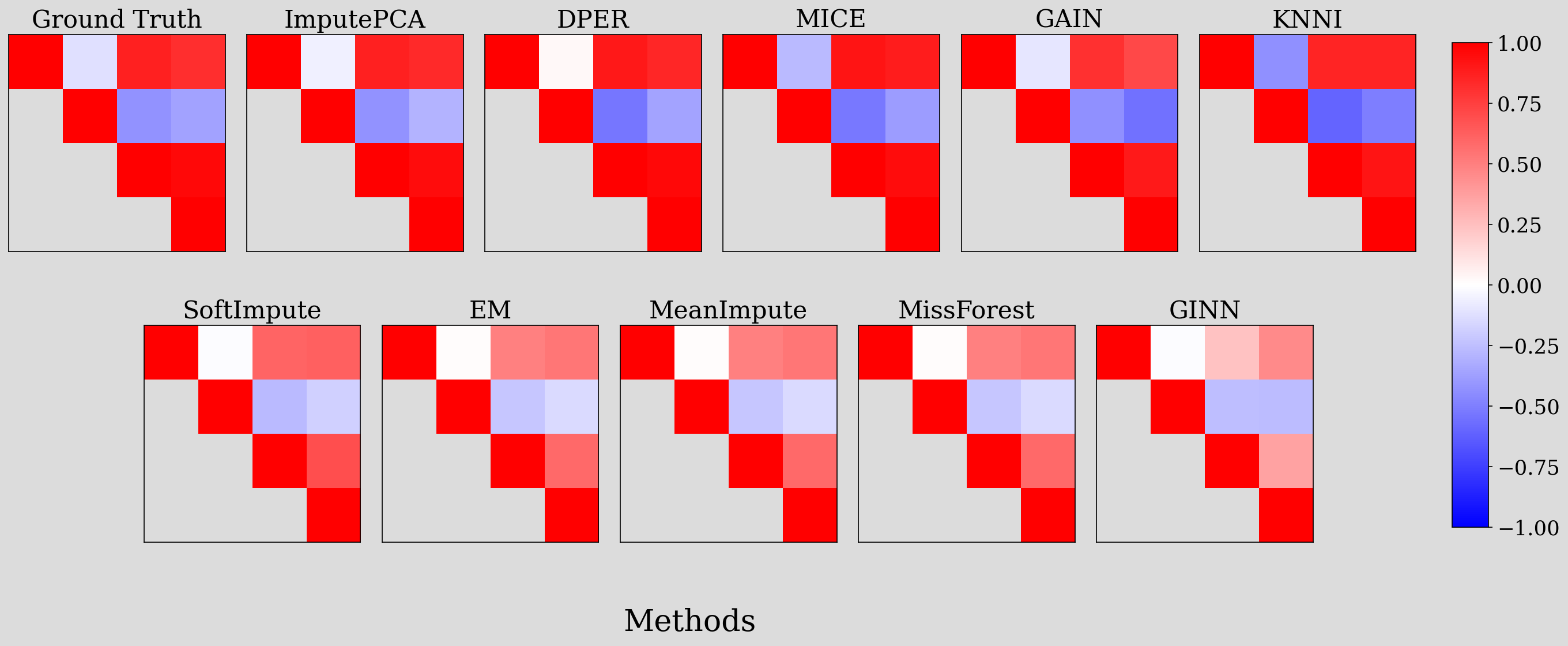}
    }\vspace{-2mm}\quad
    \subfigure[Local RMSE Difference Heatmaps for Correlation]
    {
        \label{subfig:iris_2}
        \includegraphics[width=0.9\textwidth]{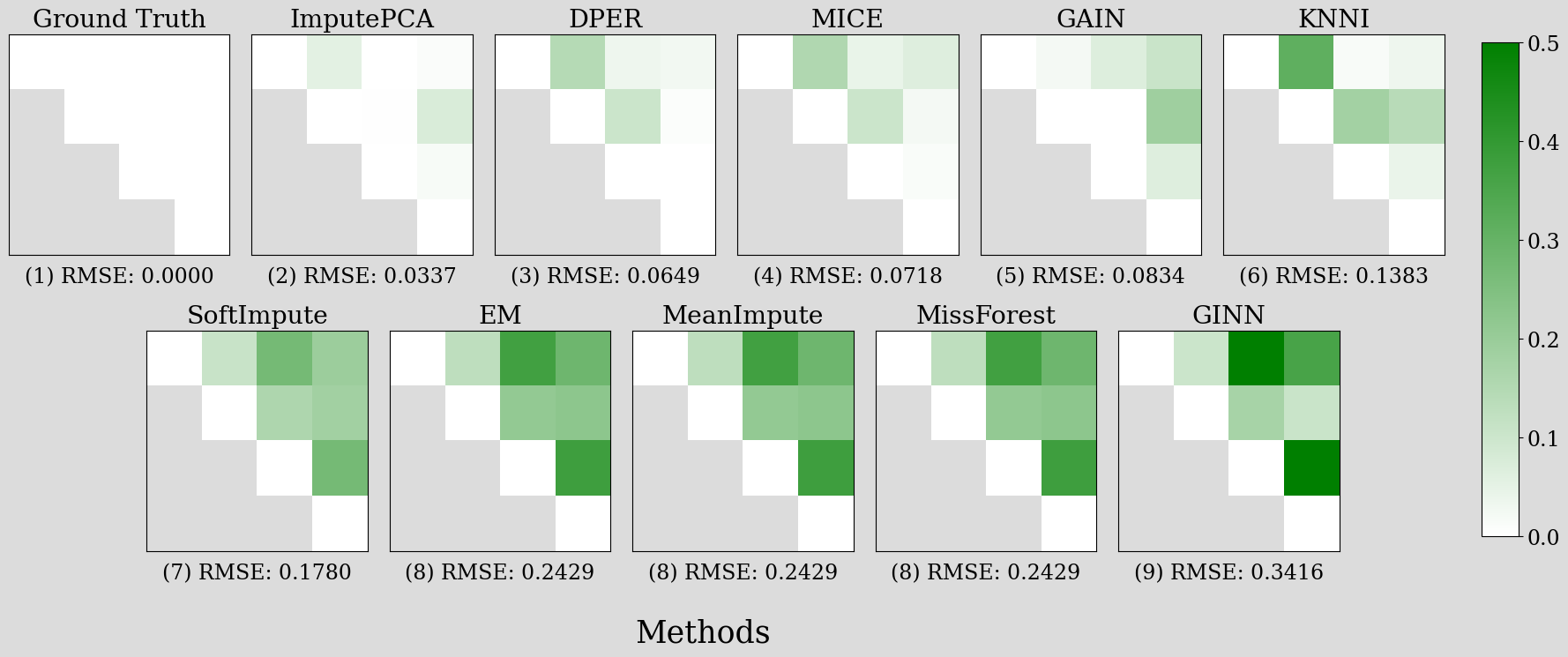}
    }\vspace{-2mm}\quad
    \subfigure[Local Difference (Matrix Subtraction) Heatmaps for Correlation]
    {
        \label{subfig:iris_3}
        \includegraphics[width=0.9\textwidth]{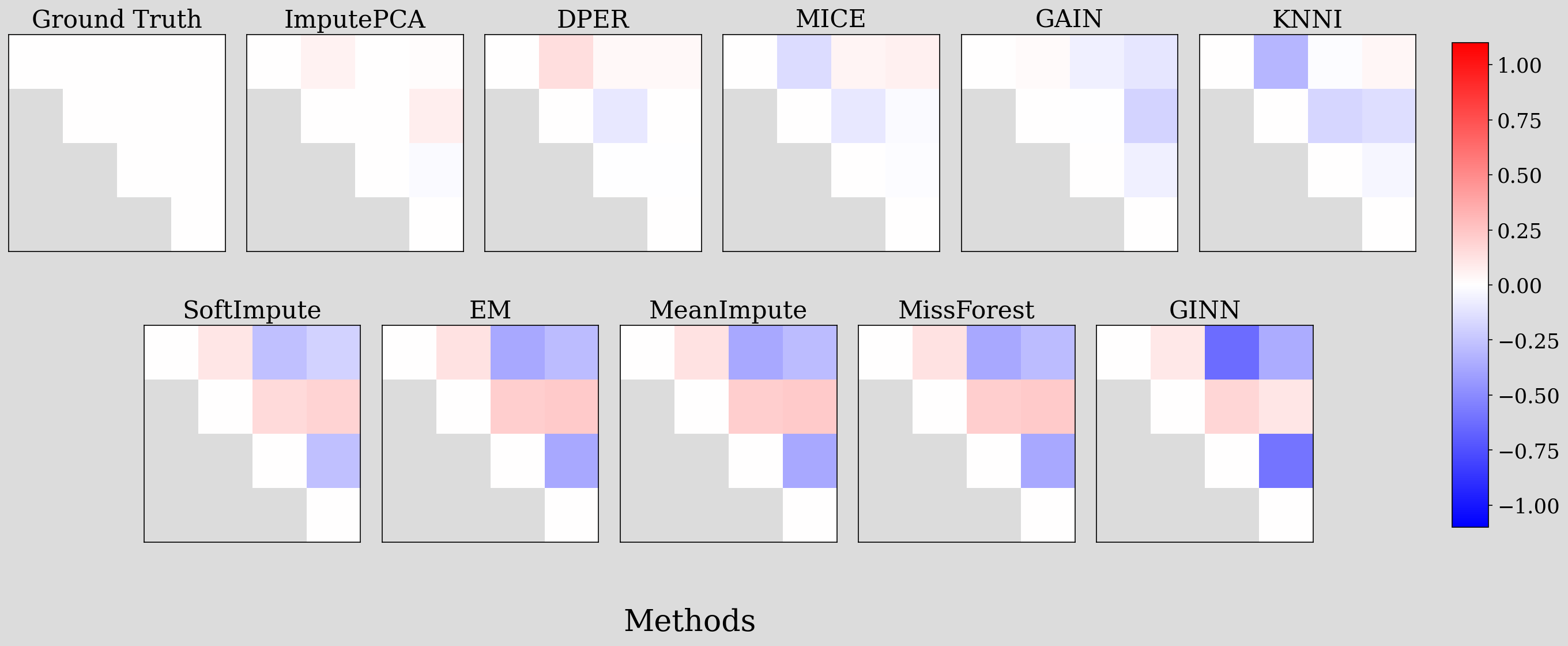}
    }
    \caption{Heatmaps for the Iris dataset at a missing rate of $50\%$}
    \label{fig:iris}
\end{figure*}

\begin{figure*}[b!]
    \centering
    \subfigure[Correlation Heatmaps]
    {
        \label{subfig:digits_all_1}
        \includegraphics[width=0.9\textwidth]{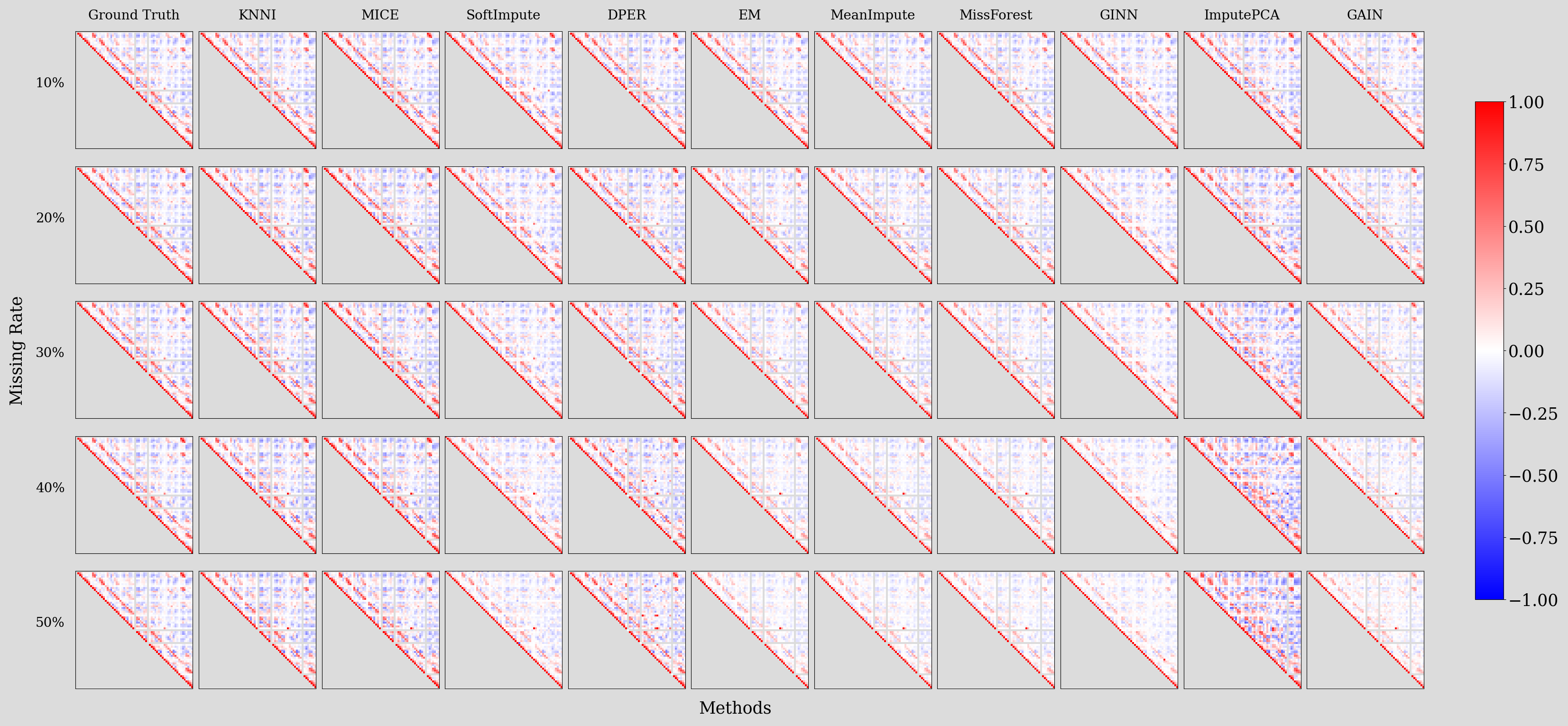}
    }\vspace{-2mm}
    \subfigure[Local RMSE Difference Heatmaps for Correlation]
    {
        \label{subfig:digits_all_2}
        \includegraphics[width=0.9\textwidth]{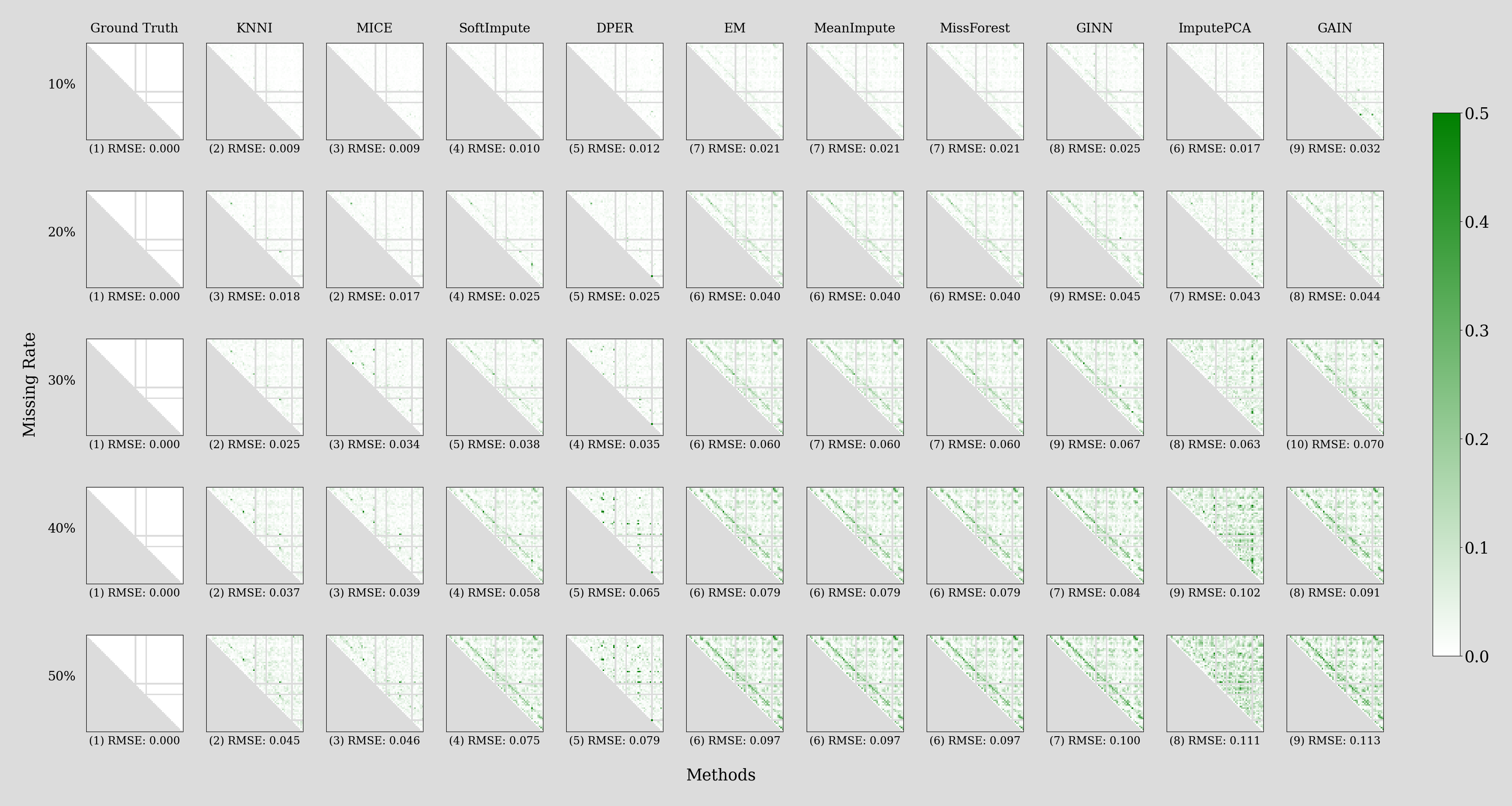}
    }\vspace{-2mm}
    \caption{Heatmaps for the Digits dataset across missing rates from $10\%$ to $50\%$}
    \label{fig:digits_all}
    \vspace{-3mm} 
\end{figure*}

\begin{figure*}[ht!]
    \centering
    \subfigure[Correlation Heatmaps]
    {
        \label{subfig:digits_1}
        \includegraphics[width=0.9\textwidth]{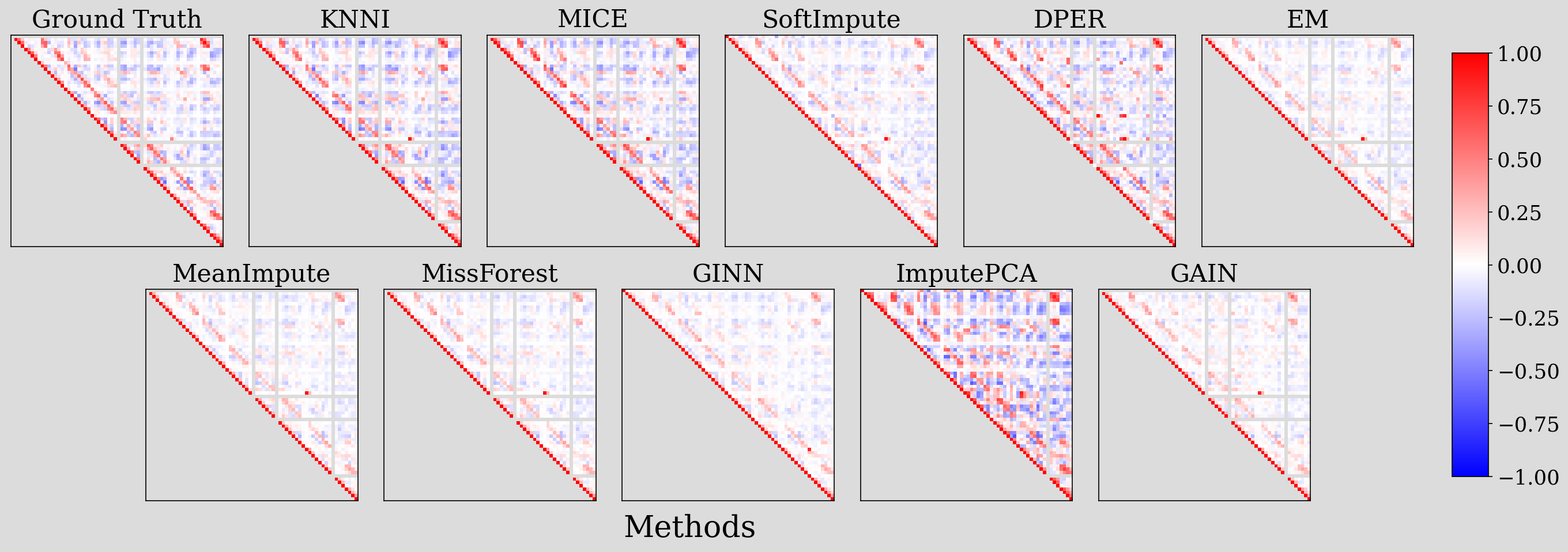}
    }\vspace{-2mm}
    \subfigure[Local RMSE Difference Heatmaps]
    {
        \label{subfig:digits_2}
        \includegraphics[width=0.9\textwidth]{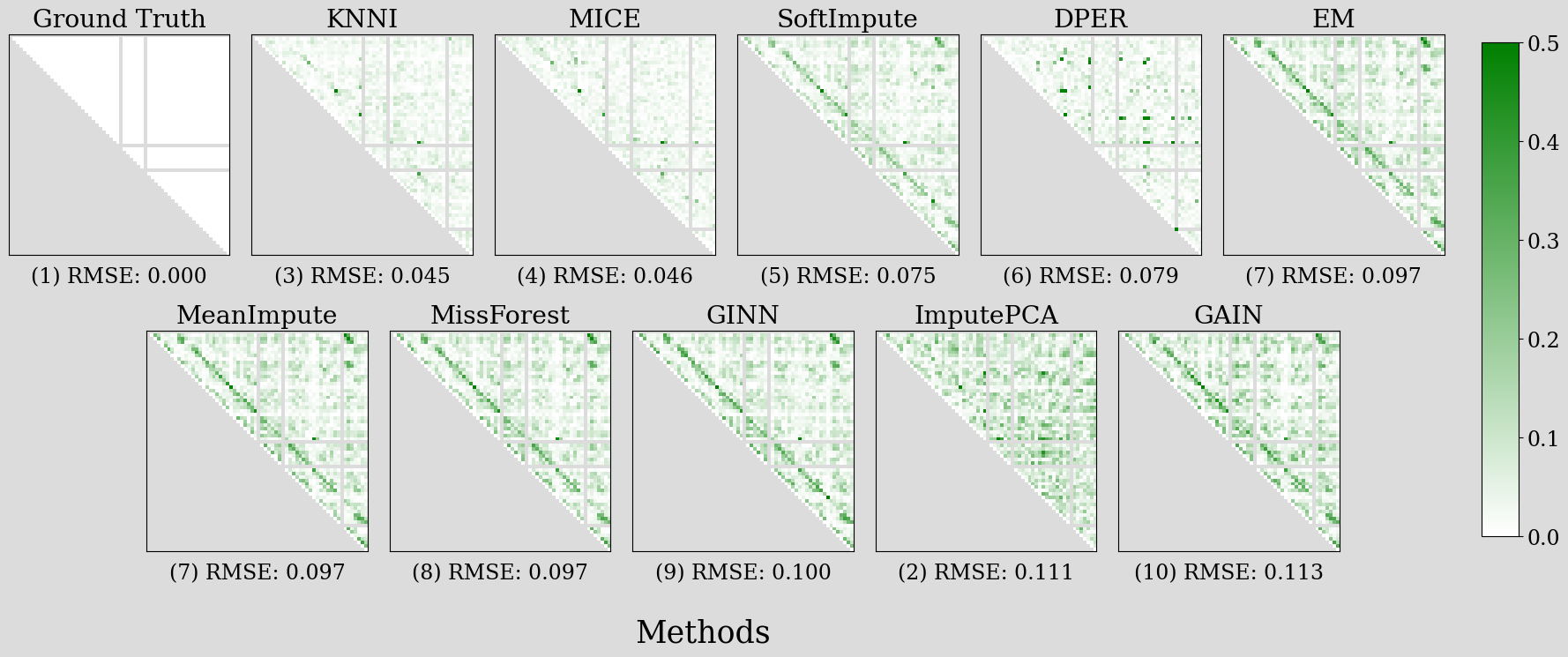}
    }\vspace{-2mm}
    \subfigure[Local Difference (Matrix Subtraction) Heatmaps for Correlation]
    {
        \label{subfig:digits_3}
        \includegraphics[width=0.9\textwidth]{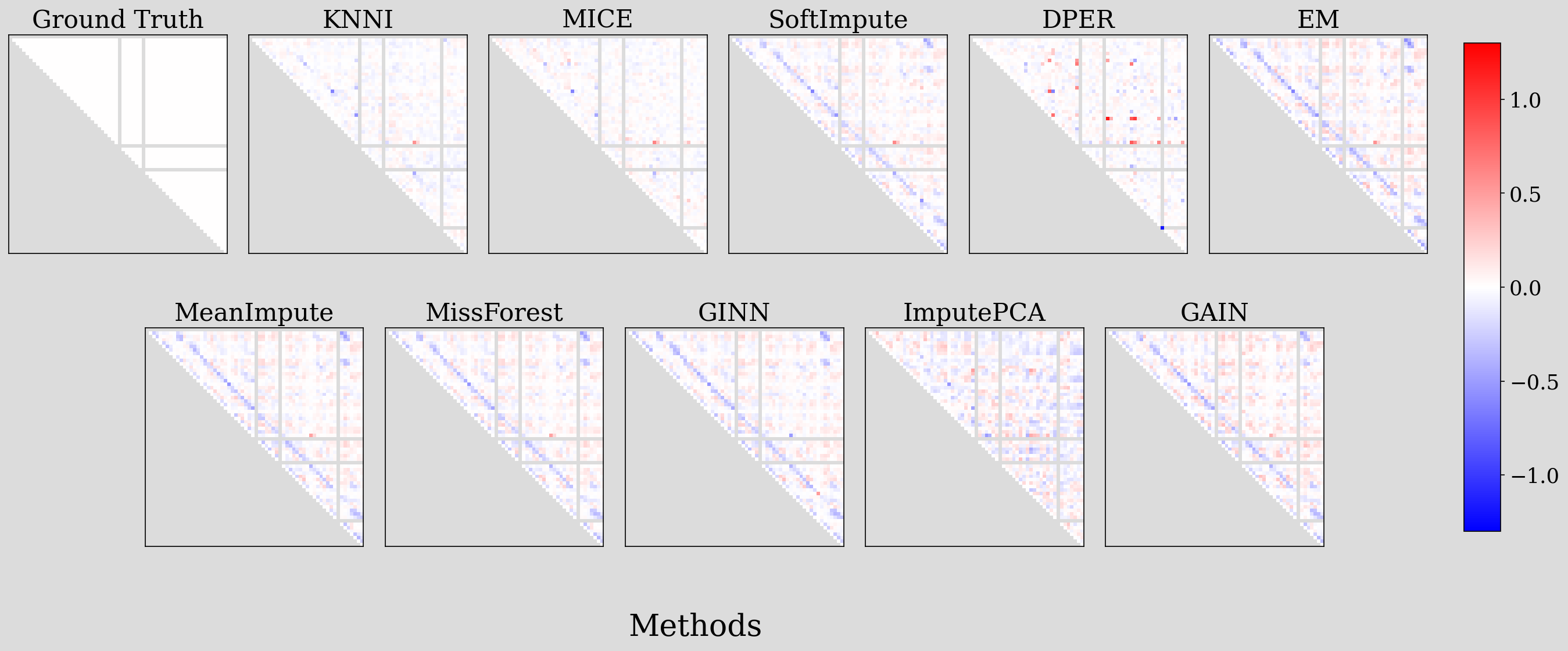}
    }\vspace{-3mm}
    \caption{Heatmaps for the Digits dataset at a missing rate of $50\%$}
    \label{fig:digits_50}
     \vspace{-2mm} 
\end{figure*}

A clear trend emerges in the RMSE values across the Iris and Digits datasets as depicted in the initial two line plots in Figure \ref{subfig:line_1}. As the missing rate increases, there is a corresponding rise in the RMSE difference. This trend is also observed in the Local RMSE Difference Heatmaps for Correlation at \ref{subfig:iris_2} \ref{subfig:digits_2}, where the color intensity of the green shades increases in both Figure \ref{subfig:iris_all_2} and Figure \ref{subfig:digits_all_2}. 
The charts also show that the RMSE rankings remain consistent by comparing the changing ranks within the Local RMSE difference heatmaps in the last plot of Figures \ref{fig:iris_all} and \ref{fig:digits_all}. This indicates that the findings about different correlation matrix estimation methods stay the same across different missing rates.

\textbf{For the Iris Dataset}, at its highest missing rate in the Local RMSE Difference Heatmaps for Correlation plot in Figure \ref{subfig:iris_2}, with Mean Imputation as the baseline, it's worth noting that EM and MissForest also exhibit RMSE values that are close to the baseline. Seven of these methods perform better than the baseline, while MissForest and GINN demonstrate higher RMSE values (0.2429 and 0.3416, respectively). Notably, DPER, MICE, and GAIN stand out, as they exhibit RMSE values lower than the median (RMSE = 0.1383, represented by KNNI). These methods are highly recommended for estimating the correlation matrix in this dataset.

The visualization of the Local Difference (Matrix Subtraction) Heatmaps for Correlation in Figure \ref{subfig:iris_3} indicates that the discrepancies are both positive and negative, represented by a mixed color between blue and white. However, we observe that the cells with an intense green color for GINN, MissForest, Mean Impute, and EM in Figure \ref{subfig:iris_2} appear blue in Figure \ref{subfig:iris_3}. This implies that features with greater divergence compared to the ground truth tend to have a negative tendency in their differences, suggesting an underestimation of the correlation between these features.

\textbf{For the Digits dataset}, in the first row of the plot layout, several methods outperform Mean Impute and have RMSE better than the median value of 0.097, namely KNN, MICE, SoftImpute, and DPER.  Interestingly, while EM has a better RMSE than Mean Impute, their RMSE values are quite similar. Like the Iris dataset, EM, Mean Impute, and MissForest exhibit nearly identical RMSE values, while GINN still displays higher RMSE values. Both ImputePCA and GAIN show larger RMSE values compared to the baseline. 

Visualizing the Local Difference (Matrix Subtraction) Heatmaps for Correlation in Figure \ref{subfig:digits_3} reveals both positive and negative disparities. An intriguing pattern emerges: certain lines running parallel to the diagonal consistently display intense green colors in Figure \ref{subfig:digits_2}, suggesting significant differences. These areas of intense green indicate that these features tend to underestimate correlation, as evidenced by the corresponding positions of these cells in Figure \ref{subfig:digits_3} appearing as blue. These positions correspond to cells with very high correlation values, as visible in Figure \ref{subfig:digits_1}. This phenomenon is consistent across SoftImpute, EM, Mean Impute, MissForest, GINN, and GAIN.

Regarding correlation estimation, methods like KNNI, MICE, SoftImpute, and DPER perform better than the median RMSE value and are thus recommended. However, practitioners should exercise caution when using SoftImpute, as it may underestimate the correlation of some highly correlated features.    

 Throughout the analysis above, techniques including KNN, MICE, SoftImpute, and DPER consistently demonstrate their effectiveness in estimating correlation matrices for the Digits dataset. Within the context of the Iris dataset, DPER, MICE, and GAIN emerge as reliable methods. Moreover, our visual observations highlight certain methods' inclination to underestimate highly correlated features in the correlation matrix within the Digits dataset.

\subsubsection{Monotone missing data} 

\begin{figure*}[t!]
    \centering
    \subfigure[Correlation Heatmaps]
    {
        \label{subfig:mnist_all_1}
        \includegraphics[width=0.9\textwidth]{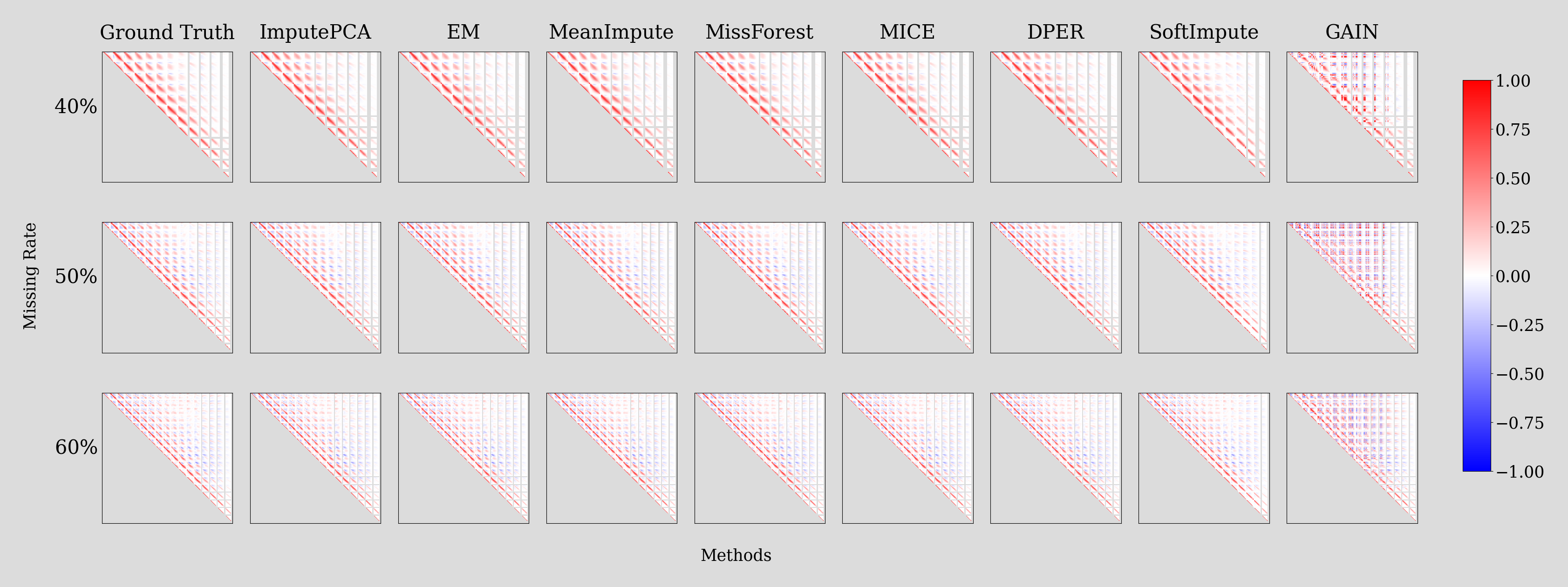}
    } 
    \vspace{-4mm} 
    \subfigure[Local RMSE Difference Heatmaps for Correlation]
    {
        \label{subfig:mnist_all_2}
        \includegraphics[width=0.9\textwidth]{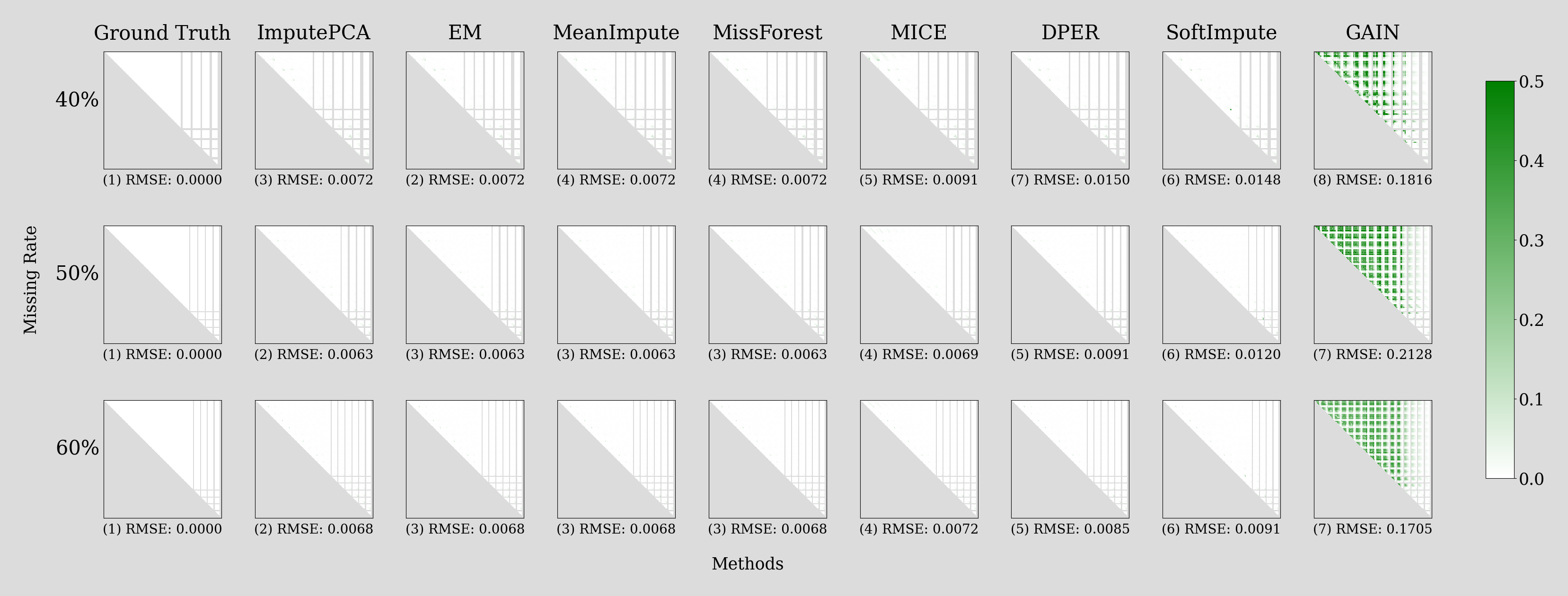}
    }
    \caption{Heatmaps for the MNIST dataset across missing rates from $40\%$ to $50\%$}
    \label{fig:mnist_all}
\end{figure*}

In contrast to the Iris and Digits datasets, the results for the plots in Figure \ref{subfig:mnist_all_2} across varying missing rates show minimal variation among methods. Most methods, except for GAIN, exhibit closely aligned RMSE values, which demonstrate the highest RMSE and are visually apparent in the RMSE plot. All of these methods, except for GAIN, are recommended for estimating the correlation matrix for this dataset. 


\begin{figure*}[ht!]
    \centering
    \subfigure[Correlation Heatmaps]
    {
        \label{subfig:mnist_1}
        \includegraphics[width=0.9\textwidth]{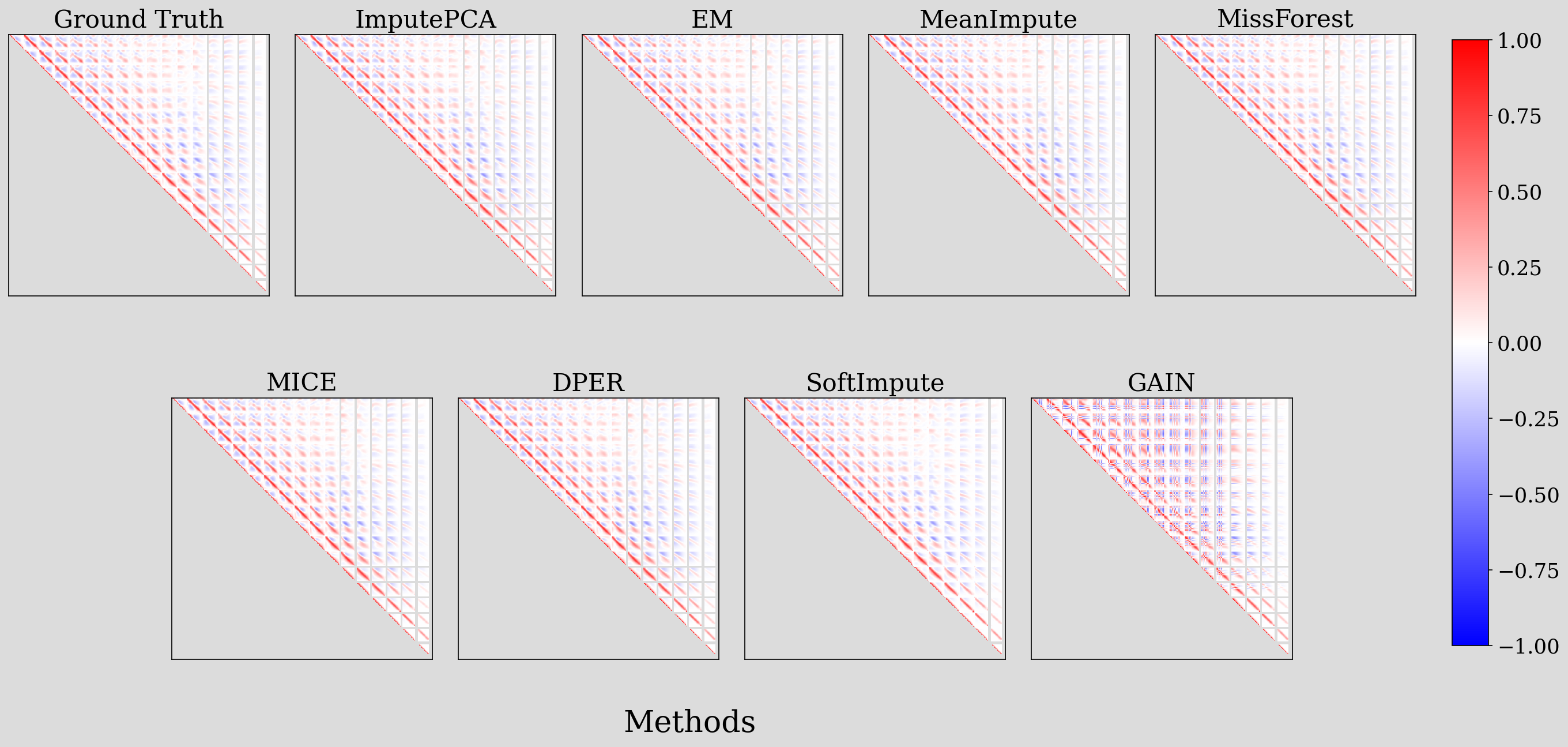}
    }\vspace{-2mm} 
    \subfigure[Local RMSE Difference Heatmaps for Correlation]
    {
        \label{subfig:mnist_2}
        \includegraphics[width=0.9\textwidth]{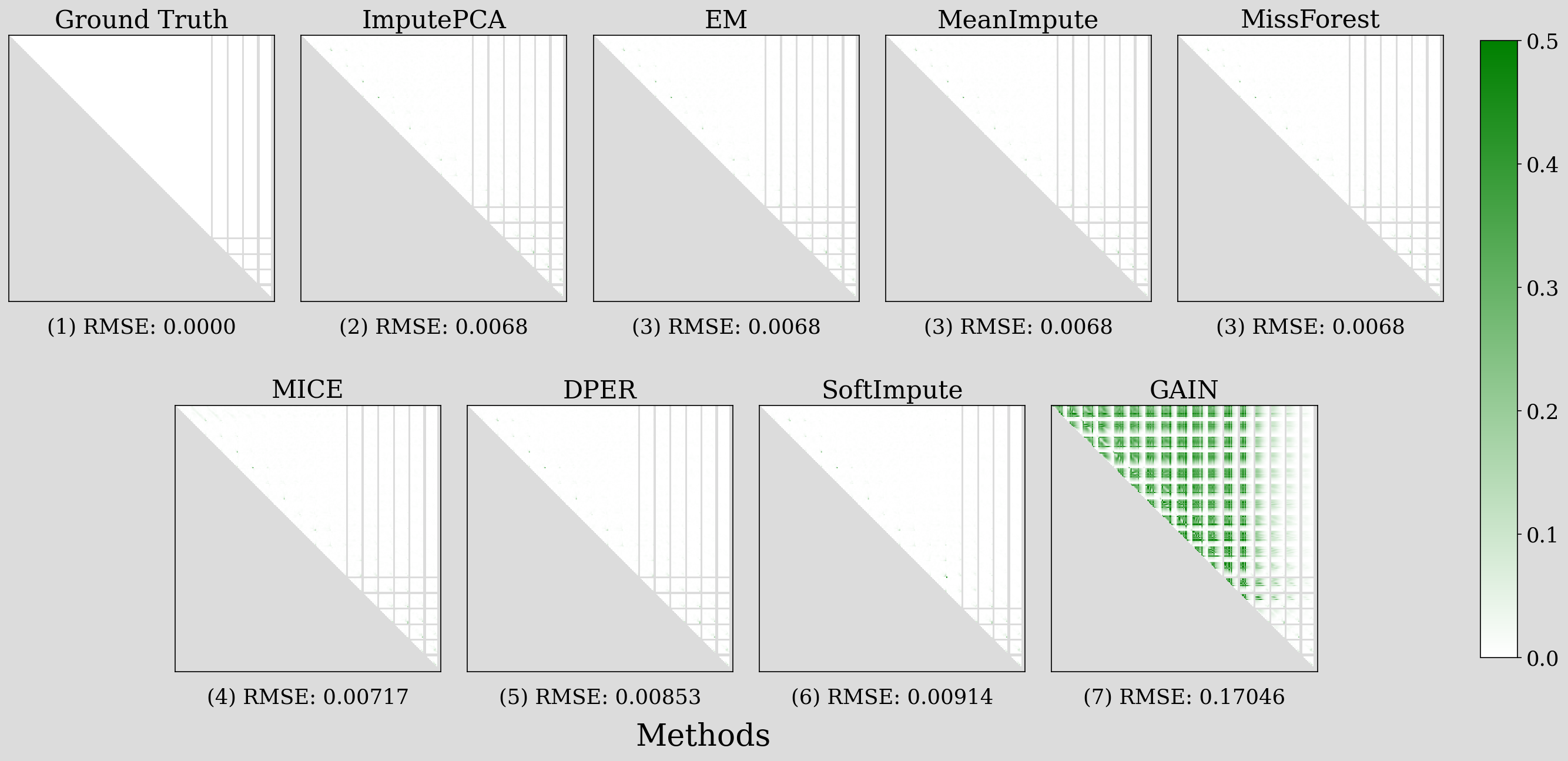}
    }\vspace{-2mm}
    \subfigure[Local Difference (Matrix Subtraction) Heatmaps  for Correlation]
    {
        \label{subfig:mnist_3}
        \includegraphics[width=0.9\textwidth]{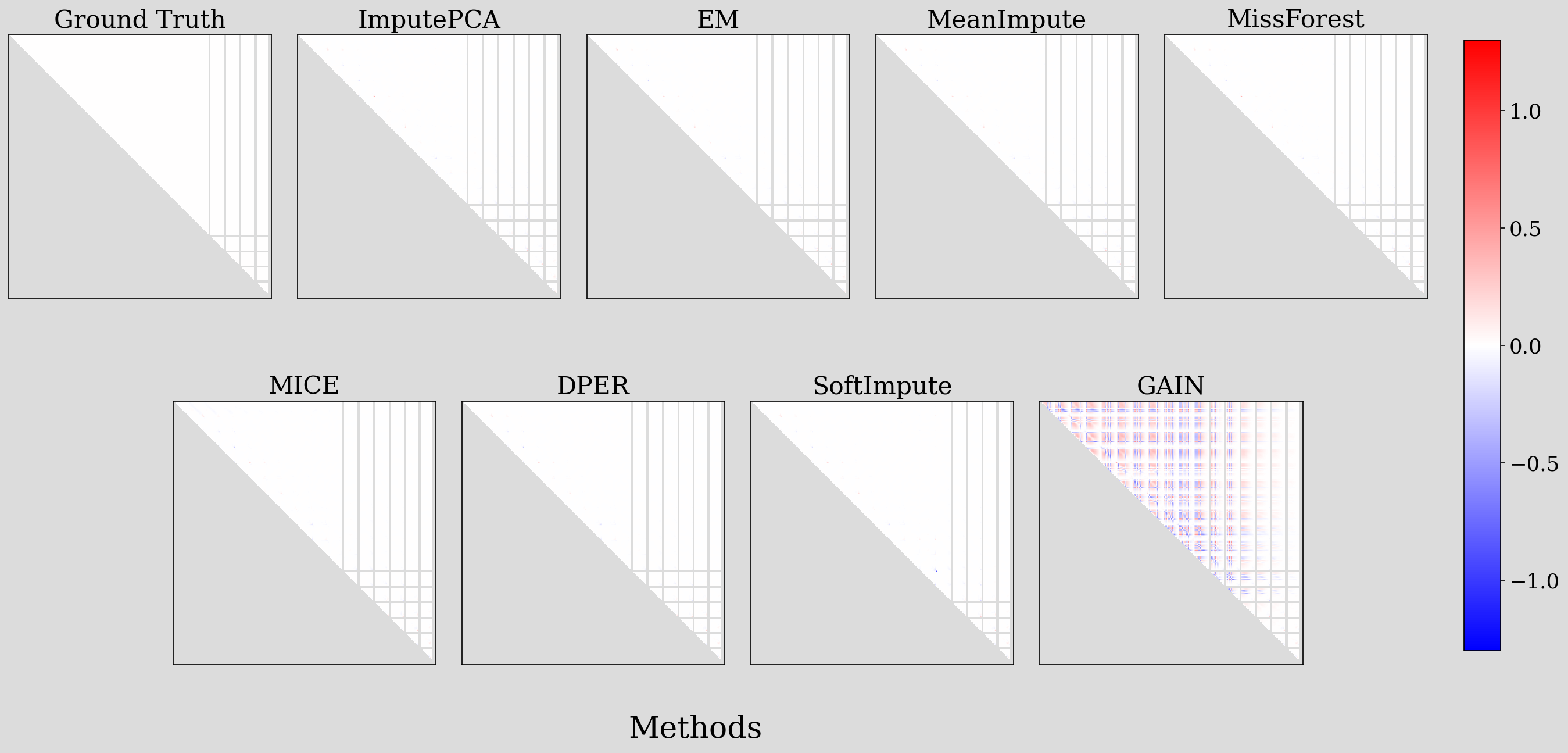}
    }\vspace{-2mm} 
    \caption{Heatmaps for the MNIST dataset at a missing rate of $50\%$}
    \label{fig:mnist}
\end{figure*}
  
\subsubsection{Analysis}
In summary, from the observations for all the datasets, we see that DPER and MICE consistently provide plots that most resemble the ground truth. 

From the experiments, one can see that a low RMSE may not imply a correlation plot reflecting the true relation between the features. For example, at a 50\% missing rate on the Digits dataset, even though softImpute and Missforest have lower RMSE than DPER, DPER still clearly resembles the ground truth plot. To understand why, suppose that we have a dataset of two features where the covariance matrix is  
$$
\mathbf{\Sigma} = \begin{pmatrix} \sigma_1 & \sigma_{12} \\ \sigma_{12} & \sigma_2\end{pmatrix}. $$ 

Then $RMSE = \sqrt{(\hat{\sigma}_1-\sigma_1)^2 + (\hat{\sigma}_2-\sigma_2)^2+2(\hat{\sigma}_{12}-\sigma_{12})^2},$
and whether $((\hat{\sigma}_1-\sigma_1)^2, (\hat{\sigma}_2-\sigma_2)^2, (\hat{\sigma}_{12}-\sigma_{12})^2)$ receives the values $(0.1, 0.5, 0.1)$ or $(0.2, 0.2, 0.2)$ will result in the same values of RMSE. 

Also, it is important to emphasize that more than relying solely on RMSE values for selecting an imputation method, and then using the imputed data based on that method to create correlation plots and draw inferences can lead to misleading conclusions. Considering Local RMSE differences in heatmaps and Local difference (matrix subtraction), heatmaps are vital to ensure precise method selection.

\section{Conclusions}  
In this work, we examined the effects of missing data on correlation plots. Crucially, we found that the techniques yielding the lowest RMSE do not consistently produce correlation plots that closely resemble those based on complete data (the ground truth). This could be because the RMSE is an average measure. Meanwhile, human eyes capture local differences in the matrix, while visual perception is better at capturing local differences, which could be better visualized by representing the local square difference on each cell of the heatmap (employed by the Local RMSE Difference Heatmaps for Correlation in our experiments). In addition, the ground truth data itself may contain noise. 
Therefore, relying solely on RMSE values for selecting an imputation method and then using the imputed data based on that method to create a correlation plot and draw inferences can lead to misleading conclusions. Considering Local RMSE differences in heatmaps and Local difference (matrix subtraction), heatmaps are vital to ensure proper inferences from correlation plots when there is missing data. 

Moreover, our analysis consistently highlights the efficiency of correlation estimation methods such as MICE and DPER across all three datasets. Notably, DPER 
is scalable and consistently produces the correlation plot that best resembles the ground truth for both small and large datasets. We hypothesize that such a performance is related to the asymptotic consistency property of DPER and will investigate this theoretically in the future.

\newpage
\begin{footnotesize}
    \bibliographystyle{unsrt}
    \bibliography{ref,ref_dimv}
\end{footnotesize}
 
\end{document}